\documentclass{article}

\usepackage[final]{corl_2020_edited} 
\usepackage{graphicx}
\graphicspath{{Figures/}}
\usepackage{wrapfig}
\usepackage{caption}
\usepackage{subcaption}
\usepackage{algorithm}
\usepackage[noend]{algpseudocode}

\usepackage{amsmath,amsfonts,bm}
\usepackage{empheq}

\def\eqref#1{equation~\ref{#1}}

\def\1{\bm{1}}

\DeclareMathAlphabet{\mathsfit}{\encodingdefault}{\sfdefault}{m}{sl}
\SetMathAlphabet{\mathsfit}{bold}{\encodingdefault}{\sfdefault}{bx}{n}

\usepackage{enumitem}
\usepackage[dvipsnames]{xcolor}

\usepackage{hyperref}
\hypersetup{
    colorlinks=true,
    linkcolor=blue,
    filecolor=magenta,      
    urlcolor=cyan,
}

\makeatletter
\algrenewcommand\ALG@beginalgorithmic{\footnotesize}
\makeatother
\algnewcommand{\LeftComment}[1][\footnotesize]{\State #1}
\algnewcommand{\BlockComment}[1][\footnotesize]{\State \textcolor{blue}{/*{#1}*/}}
\algrenewcommand\algorithmiccomment[1]{\hfill \textcolor{blue}{#1}}

\begin{document}

\begin{center}

\title{Dexterous Manipulation Primitives for the Real Robot Challenge}

\end{center}

\author{Claire Chen \thanks{Stanford University, Palo Alto, CA} \\ 
\And Krishnan Srinivasan \footnotemark[1] \\ 
\And Jeffrey Zhang \footnotemark[1] \\
\And Junwu Zhang \footnotemark[1] \\}


\maketitle


\centerline{Advised by: Lin Shao\footnotemark[1], Shenli Yuan\footnotemark[1], Preston Culbertson\footnotemark[1], Hongkai Dai \footnote{Toyota Research Institute, Los Altos, CA}, Mac Schwager\footnotemark[1], Jeannette Bohg \footnotemark[1]}

\begin{abstract}
    This report describes our approach for Phase 3 of the Real Robot Challenge. To solve cuboid manipulation tasks of varying difficulty, we decompose each task into the following primitives: moving the fingers to the cuboid to grasp it, turning it on the table to minimize orientation error, and re-positioning it to the goal position. We use model-based trajectory optimization and control to plan and execute these primitives. These grasping, turning, and re-positioning primitives are sequenced with a state-machine that determines which primitive to execute given the current object state and goal. Our method shows robust performance over multiple runs with randomized initial and goal positions. With this approach, our team placed second in the challenge, under the anonymous name ``sombertortoise'' on the leaderboard. Example runs of our method solving each of the four levels can be seen \href{https://www.youtube.com/watch?v=I65Kwu9PGmg&list=PLt9QxrtaftrHGXcp4Oh8-s_OnQnBnLtei&index=1}{in this video}.
\end{abstract}
%


\section{Real Robot Challenge Overview}

\begin{figure}[h]
    \centering
    \includegraphics[width=0.35\linewidth]{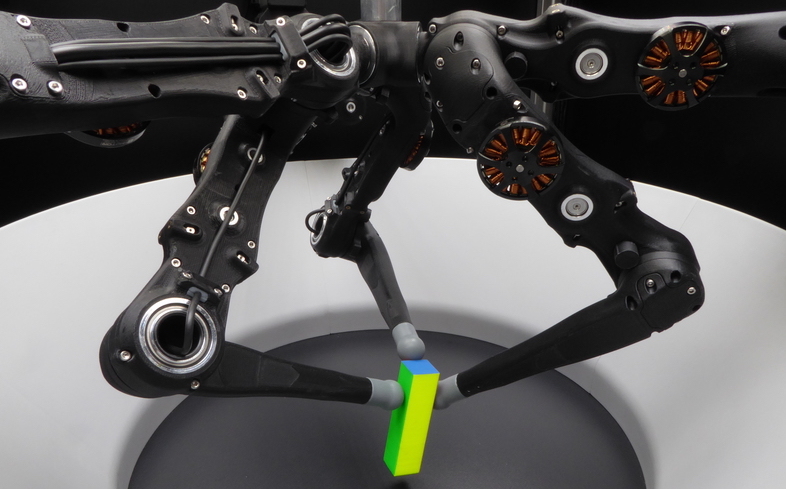}
    \caption{The Phase 3 setup consists of the TriFinger robot and a cuboid object.}
    \label{fig:robot_phase_3}
\end{figure}

The Real Robot Challenge \cite{RealRobotChallenge} invited teams to design and implement methods for performing dexterous manipulation tasks using the TriFinger robotic platform \cite{wuthrich2020trifinger}. The TriFinger robot has three fingers, each with three degrees of freedom. Phases 1 and 2 of the challenge involve manipulating a cube with a 6cm side length, first in simulation and then with the real robot. Phase 3 involves manipulating a 2cm x 2cm x 8cm cuboid object. Each phase is divided into the following 4 levels of increasing difficulty:

\begin{itemize}
    \item Level 1: Move object to randomly sampled goal position on the table. Orientation not considered.
    \item Level 2: Move object to a fixed goal position above the table. Orientation not considered.
    \item Level 3: Move object to randomly sampled goal position. Orientation not considered.
    \item Level 4: Move object to randomly sampled goal pose where both position and orientation are considered.
\end{itemize}

In all phases, we have access to robot proprioception including joint angles and fingertip forces, as well as an estimate of the object pose obtained from three cameras. We are also given full information about the shape and mass of the object, as well as the mass properties of the robot.

\section{Method}

Our approach rests on the idea that in-hand manipulation tasks can be broken down into manipulation primitives \cite{okamura2000overview}. We decompose the task of moving the object to a goal pose into three primitives:
\begin{itemize}
    \item \textit{grasp}: Move fingertips from initial positions to pre-defined contact points on cuboid. Before executing the \textit{re-position} and \textit{turn} primitives, the state machine must choose the \textit{grasp} primitive.
    \item \textit{re-position}: With three fingers on the cuboid, translate and lift to goal position.
    \item \textit{turn}: With three fingers on the cuboid, rotate it on the table and return fingers to initial positions above object.
\end{itemize}
We use model-based trajectory optimization and control to plan and execute these primitives, the details of which are described in the following sub-sections. We design a state machine to sequence the primitives. At the start of a task, the state machine selects the \textit{grasp} primitive to bring the fingers to desired contact points on the cuboid. Once the cuboid has been grasped, the state machine determines whether to \textit{re-position} or \textit{turn} the cuboid. For Levels 1-3, the state machine always chooses the \textit{re-position} primitive, as orientation error is not considered in the score. For Level 4, the state machine chooses to \textit{turn} the object to align its yaw rotation with that of the goal pose before choosing the \textit{re-position} primitive to move it to the goal position.

To grasp the object, we plan trajectories for each fingertip to reach desired contact points using trajectory optimization and track these trajectories using the simplified impedance controller from \cite{wuthrich2020trifinger}. To re-position and turn the object, we first compute a trajectory for the object pose assuming that contact forces are applied at the pre-defined contact points from the \textit{grasping} primitive, and then use this trajectory to compute the corresponding fingertip trajectories.

In addition to assuming that the fingers remain fixed at the pre-defined contact points, we also assume that we have access to prior knowledge of robot kinematics and dynamics, physical parameters of the object, as well as accurate state feedback on object pose and finger joints. Given these assumptions, a model-based trajectory optimizer and simplified impedance controller to track fingertip trajectories are sufficient for executing all of the primitives.


\begin{figure}[htbp]
    \centering
    \includegraphics[width=0.6\linewidth]{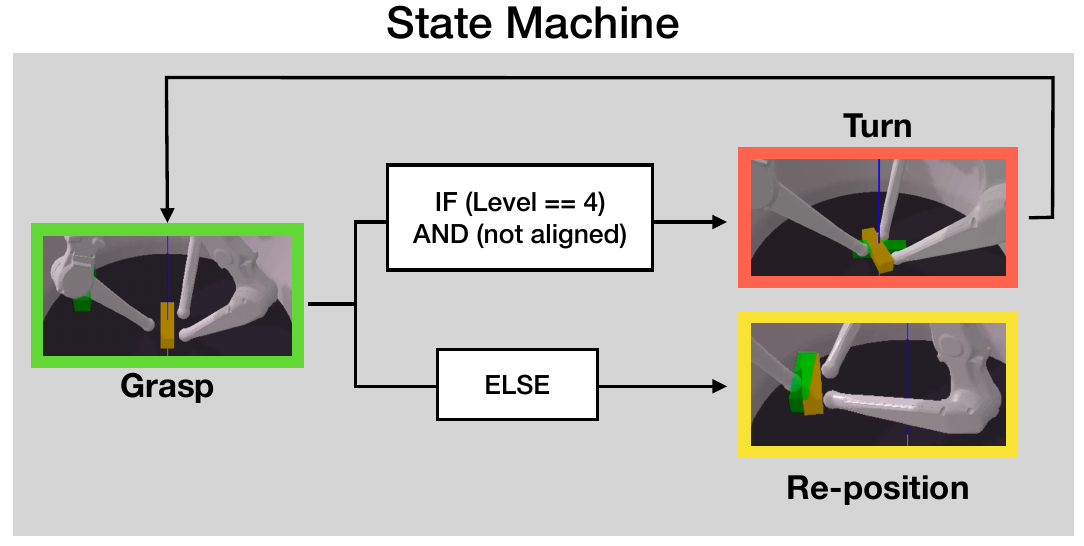}
    \caption{A diagram of the state machine. A primitive is selected based on the required level of difficulty and the difference between the object's current and goal pose. The state machine begins with the \textit{grasp} primitive to move the fingers to contact points on the object. For Levels 1-3, the state machine always chooses the \textit{re-position} primitive, as orientation error is not considered in the score. For Level 4, the state machine chooses to \textit{turn} the object to align the yaw rotation before choosing the \textit{re-position} primitive to move it to the goal position. The \textit{re-position} primitive is always assumed to be successful and is thus the terminal state.}
    \label{fig:intro}
\end{figure}


\subsection{Grasping}
To grasp the cuboid securely, we use a three-fingered grasp to pinch the cuboid across its short axis, as shown in Figure \ref{fig:intro}. This grasp is used for both the \textit{re-position} and \textit{turn} primitives. We pre-define the contact points that compose this grasp, with one finger placed on a long face of the cuboid and the other two fingers placed the opposite long face of the cuboid, equidistant from the center of the face. For each finger, we choose which face to contact based on the face that is closest to that finger. In our tests, we find that this simple face assignment strategy is enough to ensure that all fingers are able to reach the contact points without collision, since the cuboid always starts near the center of the arena. Since the object is very narrow, we perform a two-stage grasping motion by first lowering the fingers to the height of the object before then moving them towards the object to pinch it. We use trajectory optimization to first plan a trajectory to lower the fingers, and then again to pinch the cuboid.

For the cube manipulation task in Phase 2, we included a collision penalty term in the trajectory optimization problem to prevent the fingertips from making unwanted contact with the cube. While this worked well for grasping the cube, we found it much more challenging to tune the collision penalty term for a slimmer object like the cuboid. Ultimately, the trajectory optimization we use for the cuboid tasks in Phase 3 does not include collision avoidance, as the two-stage grasping motion was more reliable for preventing unwanted collisions. We acknowledge that without the need for collision avoidance, an operational space controller could have been used instead of trajectory optimization for the simple task of moving the fingertips to desired goal positions; however, we already had the trajectory optimization method implemented.

\paragraph{Fingertip position trajectory optimization:}
We formulate a direct collocation trajectory optimization problem to compute collision-free fingertip trajectories to the desired goal positions in Cartesian space. The optimization problem, shown below, considers only the kinematics of the fingers and assumes that the pose of the cuboid does not change. We find trajectories of length $T$ for joint angles $q$ and joint velocities $\dot{q}$ and use forward kinematics to compute the corresponding fingertip positions $x$ and velocities.

\begin{equation}
\begin{aligned}
& \underset{q, \dot{q}, \alpha}{\text{minimize}}\;\;\;
\sum_{t=0}^{T} (x_{\text{goal}} - x_t)^T Q (x_{\text{goal}} - x_t)
+ \sum_{t=0}^{T-1} \dot{q}^T R \dot{q}
+ \sum \alpha\\
& \text{subject to:}
\end{aligned}
\vspace*{-\baselineskip}
\end{equation}
\begin{alignat}{2}
q_{k+1} - q_k &= \frac{1}{2} (t_{k+1} - t_k) (\dot{q}_{k+1} + \dot{q}_k) & \;\;\;\forall k \in (0, 1, 2,... ,T-1) \label{eq:ft_con_kin}\\
||x_k||_2 &< r_{\text{arena}} & \;\;\;\forall k \in (0, 1, 2,... ,T-1) \label{eq:ft_con_arena}\\
(x_{\text{goal}} - x_T) ^ 2 &> \alpha\label{eq:ft_con_goal}\\
\alpha &> 0 \label{eq:ft_con_alpha}\\
q &\in [q_{\text{min}}, q_{\text{max}}]\\
\dot{q} &\in [\dot{q}_{\text{min}}, \dot{q}_{\text{max}}]
\end{alignat}

The cost function includes the distance between the fingertip positions and goal positions at each time step weighted with weight matrix $Q$, joint velocities at each time step weighted with weight matrix $R$, and the sum of all the slack variables $\alpha$. The weight matrices were tuned to produce reasonable-looking solution trajectories for the fingers. Equation \ref{eq:ft_con_kin} enforces joint velocity constraints, Equation \ref{eq:ft_con_arena} ensures that the fingertips remain within the arena radius, and Equations \ref{eq:ft_con_goal} and \ref{eq:ft_con_alpha} constrain the final fingertip positions to be equal to the goal fingertip positions, relaxed with slack variables $\alpha$. We use IPOPT \cite{ipopt} to solve all trajectory optimization problems.

\subsection{Turning and re-positioning}
Given an initial and goal pose for the object's center of mass (CoM), we use direct collocation to plan a trajectory for the object, as well as the associated forces that need to be applied at fixed contact points to move the object along this trajectory. The contact points are assumed to be fixed at the pre-defined positions specified in the \textit{grasp} primitive.

\paragraph{Object CoM trajectory optimization:}
The following optimization problem finds trajectories of length $T$ for object pose $o$, object velocity $\dot{o}$, and contact forces $\lambda^{cf}$ expressed in the corresponding local contact point reference frames $cf$.  
\begin{equation}
\begin{aligned}
& \underset{o, \dot{o}, \lambda^{cf}, \alpha}{\text{minimize}}\;\;\;
\sum_{t=0}^{T} (o_{\text{goal}} - o_t)^T Q (o_{\text{goal}} - o_t)
+ \sum_{t=0}^{T-1} (\lambda^{cf}_{\text{target}}-\lambda^{cf}_t)^T R (\lambda^{cf}_{\text{target}}-\lambda^{cf}_t)
+ \sum \alpha\\
& \text{subject to:}
\end{aligned}
\vspace*{-\baselineskip}
\end{equation}
\begin{alignat}{2}
o_{k+1} - o_k &= \frac{1}{2} (t_{k+1} - t_k) (\dot{o}_{k+1} + \dot{o}_k) & \;\;\;\forall k \in (0, 1, 2,... ,T-1) \label{eq:obj_con_dyn1}\\
\dot{o}_{k+1} - \dot{o}_k &= \frac{1}{2} (t_{k+1} - t_k) (\ddot{o}_{k+1} + \ddot{o}_k) & \;\;\;\forall k \in (0, 1, 2,... ,T-1) \label{eq:obj_con_dyn2}\\
\text{where:} \;\;\;\;\;\; & \Ddot{o} = M_{\text{obj}}^{-1} (G \lambda^{cf} + g_{\text{obj}}) \label{eq:obj_dyn}
\end{alignat}
\begin{alignat}{2}
(o_{\text{goal}} - o_T) ^ 2 &> \alpha \label{eq:obj_con_goal}\\
\alpha &> 0 \label{eq:obj_con_alpha}
\end{alignat}

We use Equation \ref{eq:obj_dyn} to define object dynamics, where $M_{\text{obj}}$ is the mass matrix of the object, $G$ is the grasp matrix, and $g_{\text{obj}}$ is the vector of gravitational forces on the object. The cost function includes the distance between the object positions and goal positions at each time step weighted with weight matrix $Q$, the difference between the contact forces and target contact forces $\lambda^{cf}_{\text{target}}$ at each time step weighted with weight matrix $R$, and the sum of all the slack variables $\alpha$. The weight matrices were tuned to produce trajectories where the object would move towards the goal pose gradually, rather than too quickly, as we found that it was most reliable to move the object slowly. The normal force components of $\lambda^{cf}_{\text{target}}$ are set to some desired normal force. We can also constrain the contact forces to lie within linear approximations of friction cones to prevent contact slippage, but in practice, we found that applying sufficient normal force was enough to prevent grasp slippage. Equations \ref{eq:obj_con_dyn1} - \ref{eq:obj_dyn} enforce the object's dynamic constraints, and Equations \ref{eq:obj_con_goal} and \ref{eq:obj_con_alpha} constrain the final object position to be equal to the goal object pose $o_{\text{goal}}$, relaxed with slack variables $\alpha$. Given the trajectories for object pose and contact forces expressed in local contact frames and given our assumption of fixed contact points, we compute the corresponding trajectories for fingertip positions $x$ and contact forces expressed in robot frame $\lambda^{rf}$.

Compared to re-positioning the object, re-orienting the object as required in Level 4 is considerably less straight-forward. While re-positioning the cuboid can be achieved with a single grasp, rotating the cuboid could potentially require re-grasping. To enable this, we introduce additional logic into our state machine. Given the difficult nature of re-orienting the object in mid-air, we only attempt to turn the object while it rests on the table. To ensure that the contact points remain reachable by each finger while rotating the cuboid, we chose a simple heuristic: we turn the object in 45 degree increments, resetting and re-grasping the object between each rotation. For example, a 120 degree rotation would be broken down into two 45 degree rotations followed by a 30 degree rotation.

\subsection{Impedance Controller}
We compute the joint torques necessary for tracking the desired fingertip trajectories in Cartesian space using the simplified impedance controller from \cite{wuthrich2020trifinger} with additional gravity compensation for the fingers (time index omitted for clarity):
\begin{equation} \label{eq:ctr2}
\tau = J^T(k_p(x_{\text{ref}}-x)+k_v(\dot{x}_{\text{ref}}-\dot{x})+\lambda^{rf}) + g_{\text{hand}}
\end{equation}
where $\tau \in\rm I\!R^9$ is the vector of joint torques to be applied to each finger, $x_{\text{ref}}$ are the desired fingertip positions from the reference trajectory, $\lambda^{rf}$ is the vector of desired contact forces to be applied by each finger in robot frame, $J$ is the Jacobian of the 3 fingers, $g_{\text{hand}}$ is the gravity compensation vector, and $k_p$ and $k_v$ are hand-tuned controller gains. 

While we were able to grasp and move the object by just taking into account feedback on fingertip positions, we also tried incorporating object pose feedback to further reduce the steady state error between current and goal object pose. We follow the method presented in \cite{wuthrich2020trifinger}, which uses a PD law to compute the wrench that needs to be applied to the object to track a desired trajectory. The noisy object orientation estimates made it difficult for us to obtain stable performance with this additional PD law, so we chose to only track fingertip trajectories.

\subsection{Reinforcement Learning}

To improve the score of the model-based method described above, we explored using several deep reinforcement learning algorithms for learning other primitives, such as pushing and turning the cuboid. We used Proximal Policy Optimization (PPO) to optimize a neural network policy $\pi_\phi$, parametrized by $\phi$, to push the object from an arbitrary starting position and orientation on the table to the center of the table with a desired goal orientation. The composite reward function integrates the objectives of minimizing position and orientation error, and is computed by the following equation:
\begin{equation}
    R(o_\text{pos}, o_\text{ori}, g_\text{pos}, g_\text{ori}) = 
      r_\text{pos}(\|o_\text{pos} - g_\text{pos}\|) + r_\text{ori}(\|o_\text{ori}^\text{yaw} - g_\text{ori}^\text{yaw}\|).
\end{equation}
The position reward function, $r_\text{pos}$, is \[
r_\text{pos}(d) = \begin{cases}
1, & \text{if $d < 0.05$} \\
\frac{1}{9 (20d)^2+1} & \text{$0.05 < d < 0.075$} \\
0 & \text{o.w.}
\end{cases}
\]
and similarly, the orientation reward function $r_\text{ori}$, is computed by \[
r_\text{ori}(d) = \begin{cases}
1, & \text{if $d < \frac{\pi}{8}$} \\
\frac{1}{9 (8^{-1}\pi d)^2+1} & \text{if $\frac{\pi}{8} < d < \frac{3\pi}{16}$} \\
0 & \text{o.w.}
\end{cases}
\]

The objective from \cite{Schulman2017ProximalPO} is then optimized to learn a policy:
\begin{equation}
    L(s, a, \phi_k, \phi) = \min \bigg( \frac{\pi_\phi(a|s)}{\pi_{\phi_k}(a|s)}
    A^{\pi_{\phi_k}}(s,a), \;
    \text{clip}\bigg( \frac{\pi_\phi(a|s)}{\pi_{\phi_k}(a|s)}, 1-\epsilon, 1+\epsilon \bigg) A^{\pi_{\phi_k}}(s,a) \bigg),
\end{equation}
where $s$ and $a$ are an arbitrary state and action, $A^{\pi_{\phi}}(s,a)$ is the advantage function, and $\phi_k$ is the policy parameters at the previous optimization step $k$. 

The resulting policy forms a new \emph{Push-Turn} primitive and pushes the object to a position that is a) easier for the grasp primitive to grasp the object from and b) avoids calling the \emph{turn} primitive, which requires more time to complete. While this policy was able to perform the rotation task in simulation, due to robustness issues when transferred to the real robot, we opted to use the optimal controller alone in the final submission for phase 3, as it showed less variance in the Level 4 task score.


\section{Results and Discussion}

\begin{table}[htbp]
\begin{center}
\begin{tabular}{c|c|c|ccc|ccc}
\textbf{}  & \textbf{Level 1} & \textbf{Level 2} & \multicolumn{3}{c|}{\textbf{Level 3}} & \multicolumn{3}{c}{\textbf{Level 4}} \\
Goal \#       & 1 & 1 & 1 & 2 & 3 & 1 & 2 & 3  
\\ \hline \rule{0pt}{2ex}
Median score & -3455 & -7889 & -12241 & -8676 & -11775 & -49773 & -26250 & -39230
\end{tabular}
\end{center}
\caption{The median scores of our method for different goals and levels. Per level and goal, we execute three runs each starting from a random initial position. For Levels 1 and 2, we show results for one randomly sampled goal position each. For Levels 3 and 4, we show results for three randomly sampled goal poses.}
\label{goals-levels-table}
\end{table}

Our current method is able to turn the cuboid on the table and re-position it to a goal pose. We evaluate our method on each level. Table \ref{goals-levels-table} shows the median scores over three runs per goal. The score for an episode is computed by summing the negated error between current and goal object pose over each time step of a two minute episode, i.e. scores closer to zero are better. As expected, the scores depend heavily on the goal pose; goal poses closer to the center of the arena, where the cuboid starts, result in better scores, as it takes fewer time steps for the object to reach the goal. For example, goals \#1 and \#2 in Level 3 are approximately 15cm and 10cm away from the center of the arena, respectively. Although the median final error to the goal position is approximately 1cm for both goals, the final scores are quite different. On the real robot, the initial pose of the cuboid is always near the center of the arena, but with some randomness due to the robot initialization procedure. As a consequence, we observe that the varied initial poses of the cuboid can cause large variations in scores across runs for the same goal pose. This was particularly true for Level 4 goals, since the initial orientations varied greatly. 

Although our method is able to complete Levels 1, 2, and 3, as well as make good progress in Level 4, there are several improvements that could be made in the future. Firstly, our method assumes accurate knowledge of object pose, and is therefore not robust to the noisy object orientation measurements from the visual tracker. Secondly, our method still suffered from steady state errors when re-positioning the cuboid, which could be improved by taking into account object pose feedback with either another feedback law or a learned policy. For instance, a hybrid control policy combining RL with impedance control in a hierarchical or residual policy could be helpful in reducing some of these errors, and aid in planning a trajectory that requires reorienting the object. Finally, more sophisticated reasoning could be used to choose grasps and sequence rotations for re-orienting the object.well


\section{Acknowledgements}
We thank the challenge organizers for providing us with the opportunity to use the TriFinger robot. We would especially like to thank Felix Widmaier for the very prompt responses to our many questions. We look forward to using the TriFinger platform for more exciting projects in the future!


\end{document}